\documentclass[twocolumn,journal]{IEEEtran}
\usepackage[T1]{fontenc}
\usepackage[latin9]{inputenc}
\usepackage{amsmath}
\usepackage{graphicx}
\usepackage[unicode=true,
 bookmarks=true,bookmarksnumbered=true,bookmarksopen=true,bookmarksopenlevel=1,
 breaklinks=false,pdfborder={0 0 0},pdfborderstyle={},backref=false,colorlinks=false]
 {hyperref}
\hypersetup{pdftitle={Your Title},
 pdfauthor={Your Name},
 pdfpagelayout=OneColumn, pdfnewwindow=true, pdfstartview=XYZ, plainpages=false}
\usepackage{breakurl}

\makeatletter

\providecommand{\tabularnewline}{\\}

 \let\oldforeign@language\foreign@language
 \DeclareRobustCommand{\foreign@language}[1]{%
   \lowercase{\oldforeign@language{#1}}}

\usepackage[caption=false,font=footnotesize]{subfig}
\usepackage[normalem]{ulem}

\makeatother

\begin{document}

\title{Improving Dermoscopic Image Segmentation with Enhanced Convolutional-Deconvolutional
Networks}

\author{Yading~Yuan{*} and Yeh-Chi~Lo\thanks{Y.~Yuan and Y-C Lo are with the Department of Radiation Oncology,
Icahn School of Medicine at Mount Sinai, New York, NY, USA, e-mail:
\protect\href{mailto:yading.yuan@mssm.edu}{yading.yuan@mssm.edu},
yeh-chi.lo@mountsinai.org.}}

\IEEEspecialpapernotice{}

\markboth{submitted to IEEE Journal of Biomedical and Health Informatics}{Yading Yuan \MakeLowercase{\emph{et al.}}: Improving Dermoscopic
Image Segmentation with Enhanced Fully Convolutional-Deconvolutional
Networks}
\maketitle
\begin{abstract}
Automatic skin lesion segmentation on dermoscopic images is an essential
step in computer-aided diagnosis of melanoma. However, this task is
challenging due to significant variations of lesion appearances across
different patients. This challenge is further exacerbated when dealing
with a large amount of image data. In this paper, we extended our
previous work by developing a deeper network architecture with smaller
kernels to enhance its discriminant capacity. In addition, we explicitly
included color information from multiple color spaces to facilitate
network training and thus to further improve the segmentation performance.
We extensively evaluated our method on the ISBI 2017 skin lesion segmentation
challenge. By training with the $2000$ challenge training images,
our method achieved an average Jaccard Index (JA) of $0.765$ on the
$600$ challenge testing images, which ranked itself in the first
place in the challenge. 
\end{abstract}


\IEEEpeerreviewmaketitle{}

\section{Introduction}

\IEEEPARstart{M}{alignant} melanoma is among the most rapidly growing
cancers in the world \cite{Siegel:2016} and dermoscopy is the most
commonly used $in$ $vivo$ imaging modality that provides a better
visualization of subsurface structures of pigmented skin lesions \cite{vestergaard2008}.
While this technique allows dermatologists to detect early stage melanoma
that are not visible by human eyes, visual interpretation alone is
a time-consuming procedure and prone to inter- and intra-observer
variabilities. Therefore, automated and accurate analysis of melanoma
has become highly desirable in assisting dermatologists for improving
their efficiency and objectivity when interpreting dermoscopic images
in clinical practice \cite{korotkov2012}.

Automatic skin lesion segmentation is an essential component in computer-aided
diagnosis (CAD) of melanoma \cite{ganster2001automated} \cite{celebi2007methodological}.
However, this is a very challenging task due to significant variations
in location, shape, size, color and texture across different patients.
In addition, some dermoscopic images have low contrast between lesion
and surrounding skin, and suffer from artifacts and intrinsic features
such as hairs, frames, blood vessels and air bubbles. Existing lesion
segmentation methods based on clustering, thresholding, region growing,
or deformable models have shown limited success in solving this difficult
problem when applying to a large amount of image data \cite{korotkov2012,silveira2009,celebi2009lesion}.

Recent development of deep learning has revolutionized the field of
machine learning and computer vision. Deep learning techniques, especially
deep convolutional neural networks \cite{lecun1998gradient}, have
been rapidly adopted in various medical image analysis problems, including
body recognition \cite{yan2016multi}, lesion detection \cite{cirecsan2013mitosis},
image registration \cite{miao2016cnn}, segmentation \cite{cha2016urinary}
and classification \cite{esteva2017dermatologist}. In particular,
Yu et al. \cite{yu2017automated} introduced a deep residual network
with more than $50$ layers for automatic skin lesion segmentation,
in which several residual blocks \cite{He_2016_CVPR} were stacked
together to increase the representative capability of their model.
In \cite{bi2017dermo}, Bi et al. proposed a multi-stage approach
to combine the outputs from multiple cascaded fully convolutional
networks (FCNs) to achieve a final skin lesion segmentation. In our
recent study \cite{yuan2017automatic}, we developed a fully automatic
method for skin lesion segmentation by leveraging a $19$-layer deep
FCN that is trained end-to-end and does not rely on prior knowledge
of the data. Furthermore, we designed a novel loss function based
on Jaccard distance that is directly related to image segmentation
task and eliminates the need of sample re-weighting. Experimental
results showed that our method outperformed other state-of-the-art
algorithms on two benchmark datasets - one is from ISBI 2016 challenge
titled as $skin$ $lesion$ $analysis$ $towards$ $melanoma$ $detection$
\cite{gutman2016skin}, and the other is the PH2 dataset \cite{mendoncca2013ph}.

In this paper, we present a major extension of our previous work to
further enhance our model in automatic skin lesion segmentation. Specifically,
1) we investigate the potential of using a deeper network architecture
with smaller convolutional kernels such that the new model has increased
discriminative capacity to handle a larger variety of image acquisition
conditions; 2) besides Red-Green-Blue (RGB) channels, we also investigate
the use of channels in other color spaces, such as Hue-Saturation-Value
(HSV) and CIELAB \cite{plataniotis2013color}, as additional inputs
to our network that aim for a more efficient training while controlling
over-fitting; 3) we evaluate the proposed framework on ISBI 2017 $Skin$
$Lesion$ $Analysis$ $Towards$ $Melanoma$ $Detection$ challenge\footnote{https://challenge.kitware.com/\#challenge/583f126bcad3a51cc66c8d9a}
datasets. Experimental results demonstrated a significant performance
gain as compared to our previous model, ranking itself as the first
place among $21$ final submissions.

\section{Dataset and Pre-processing}

We solely used ISBI 2017 challenge datasets for training and validating
the proposed deep learning model named as convolutional-deconvolutional
neural network (CDNN). As compared to ISBI 2016 challenge, the image
database in 2017 is doubled in size and includes a larger variety
of tumor appearance on dermoscopic images, which makes the automatic
lesion segmentation much more challenging. Specifically, the 2017
database includes a training dataset with $2000$ annotated dermoscopic
images ($374$ melanomas), and a blind held-out testing dataset with
$600$ images ($117$ melanomas). In order to facilitate model validation,
a small independent dataset with $150$ images ($30$ melanomas) is
publicly available for challenge participants to fine tune the hyper-parameters
before submitting the final results on the testing dataset. The image
size ranges from $540\times722$ to $4499\times6748$. By observing
most of the images in the training set have a height to width ratio
of $3:4$, we resized all the images to $192\times256$ using bi-linear
interpolation to keep balance between segmentation performance and
computation cost.

\begin{figure}[tp]
\begin{centering}
\textsf{\includegraphics[scale=0.5]{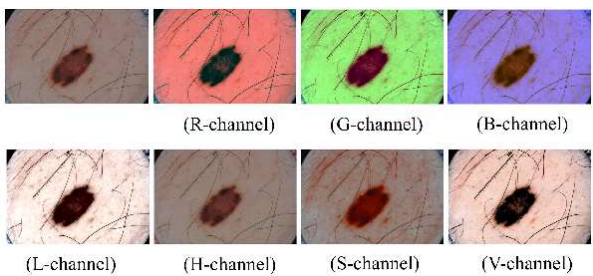}}
\par\end{centering}
\caption{The appearance of a dermoscopic image with color manipulation. The
left image in the first row shows the original image, and the rest
are the adjusted images by normalizing the contrast of each channel
to $[5,\,95]$ percentile window, respectively.\label{fig:color-adjustment}}
\end{figure}

While RGB is the most popular space to represent color information
of natural images, its usage in image segmentation is however restricted
by the fact that the RGB channels are not independent to each other.
Thus, when applying image augmentation onto each channel during network
training, the resultant image may have unrealistic appearance that
makes the training less efficient. In order to address this issue
and to use color information more effectively, we added the three
channels from HSV color space and the lightness channel (L) from CIELAB
color space, which separate $luma$ (image intensity) from $chroma$
(color information) to allow independent process on these two types
of information. Figure \ref{fig:color-adjustment} shows how a dermoscopic
image can be adjusted by normalizing the contrast of each channel
to $[5,\,95]$ percentile window. No other pre-processing was performed,
so the input dimension to our CDNN model is $192\times256\times7$.

\section{Methods}

The CDNN model we proposed belongs to FCNs category, which allows
an entire image segmentation in a single pass instead of classifying
the central pixel of a small image patch each time. Using an entire
image as input not only makes model training and inference more efficient
, but also includes much richer contextual information as compared
to small patches, which usually leads to more robust and accurate
segmentation.

\subsection{CDNN model}

In our previous work \cite{yuan2017automatic}, we proposed a CDNN
model containing $19$ layer with $290,129$ trainable parameters.
This model consists of two pathways, in which the convolutional path
resembles a traditional CNN that extracts a hierarchy of image features
from low to high complexity, and the deconvolutional path transforms
the aggregated features and reconstructs the segmentation map from
coarse to fine resolution. In this way, CDNN can take both global
information and fine details into account for tumor segmentation.

In order to improve the discriminative power of CDNN to handle a much
larger variety of image acquisition conditions, we extend the previous
model in the following two aspects. Firstly, we employ small $3\times3$
kernels in nearly all the convolutional and de-convolutional layers
and maintain the same effective receptive field size of larger kernels
by adding additional layers. This deeper architecture reduces the
number of weights while allowing more non-linear transformations on
the data. Then we increase the number of features in each layer in
order to boost the capacity of the CDNN model. All convolutional and
de-convolutional layers use Rectified Linear Units (ReLUs) as the
nonlinear activation function \cite{krizhevsky2012imagenet}, and
batch normalization \cite{ioffe2015batch} is applied to reduce the
internal covariate shift for each training mini-batch . The batch
size was set as $18$ in this study. Eventually the network contains
$29$ layers with $5,042,589$ trainable parameters. The filter size
and number of channels of the output feature maps are shown in Table
\ref{tab:cdnn}.

\begin{table}[tbh]
\centering

\caption{Architectural details of the proposed CDNN model (Abbrevations: conv:
convolutional layer; pool: max-pooling layer; decv: deconvolutional
layer, ups: upsampling layer). \label{tab:cdnn}}

\begin{tabular}{cccccc}
 &  &  &  &  & \tabularnewline
\hline 
\hline 
\noalign{\vskip2mm}
Conv & Filter & Features & Deconv & Filter & Features\tabularnewline[2mm]
\hline 
conv-1-1 & $3\times3$ & $16$ & decv-1 & $3\times3$ & $256$\tabularnewline
conv-1-2 & $3\times3$ & $32$ & ups-1 & $2\times2$ & $256$\tabularnewline
pool-1 & $2\times2$ & $32$ & decv-2-1 & $3\times3$ & $256$\tabularnewline
conv-2-1 & $3\times3$ & $64$ & decv-2-2 & $3\times3$ & $128$\tabularnewline
conv-2-2 & $3\times3$ & $64$ & ups-2 & $2\times2$ & $128$\tabularnewline
pool-2 & $2\times2$ & $64$ & decv-3-1 & $4\times4$ & $128$\tabularnewline
conv-3-1 & $3\times3$ & $128$ & decv-3-2 & $3\times3$ & $128$\tabularnewline
conv-3-2 & $4\times4$ & $128$ & ups-3 & $2\times2$ & $128$\tabularnewline
pool-3 & $2\times2$ & $128$ & decv-4-1 & $3\times3$ & $64$\tabularnewline
conv-4-1 & $3\times3$ & $256$ & decv-4-2 & $3\times3$ & $32$\tabularnewline
conv-4-2 & $3\times3$ & $256$ & ups-4 & $2\times2$ & $32$\tabularnewline
pool-4 & $2\times2$ & $256$ & decv-5-1 & $3\times3$ & $16$\tabularnewline
conv-5 & $3\times3$ & $512$ & output & $3\times3$ & $1$\tabularnewline
\hline 
\hline 
 &  &  &  &  & \tabularnewline
\end{tabular}
\end{table}

The CDNN model represents an end-to-end mapping from the input image
to a segmentation map, where each element is the probability that
the corresponding input pixel belongs to the tumor. Those trainable
parameters are learned from training data by minimizing a loss function.
We use a loss function based on Jaccard distance \cite{yuan2017automatic}:
\begin{equation}
L_{d_{J}}=1-\frac{\underset{i,j}{\sum}(t_{ij}p_{ij})}{\underset{i,j}{\sum}t_{ij}^{2}+\underset{i,j}{\sum}p_{ij}^{2}-\underset{i,j}{\sum}(t_{ij}p_{ij})},\label{eq:ja-loss}
\end{equation}
where $t_{ij}$ and $p_{ij}$ are target and the output of pixel $(i,\,j)$,
respectively. As compared to the conventionally used cross-entropy,
the proposed loss function is directly related to image segmentation
task because Jaccard index is a common metric to assess medical image
segmentation accuracy, especially in this ISBI 2017 challenge. Meanwhile,
this loss function is well adapted to the problems with high imbalance
between foreground and background classes as it doesn't require any
class re-balancing.

Given a new test dermoscopic image, it is firstly rescaled to $192\times256$
with $7$ color channels and the CDNN model is applied to yield a
segmentation map. A dual-thresholds method is then developed to generate
a binary tumor mask from the CDNN output. In this method, a relatively
high threshold ($th_{H}=0.8$) is firstly applied to determine the
tumor center, which is calculated as the centroid of the region that
has the largest mass among the candidates from thresholding. Then
a lower threshold, $th_{L}=0.5$, is applied to the segmentation map.
After filling small holes with morphological dilation, the final tumor
mask is determined as the region that embraces the tumor center. 

\subsection{Implementation details}

Our CDNN model is implemented with Python using publicly available
Theano \cite{team2016theano} and Lasagne \footnote{http://github.com/Lasagne/Lasagne}
packages. The model was trained from scratch using Adam stochastic
optimization method \cite{kingma2014adam} that adaptively adjusts
the learning rate based on the first and the second-order moments
of the gradient at each iteration. The initial learning rate $\alpha$
was set as $0.003$. 

In order to reduce overfitting, we add two dropout layers with $p=0.5$
before conv-4-1 and decv-5-1 in Table \ref{tab:cdnn}. In addition,
we implement two types of image augmentation to further improve the
robustness of the proposed model under a wide variety of image acquisition
conditions. One consists of a series of geometric transformations,
including randomly flipping, shifting, rotating as well as scaling.
The other type focuses on randomly normalizing the contrast of each
channels in the training images. Note that these augmentations only
require little extra computation, so the transformed images are generated
from the original images for every mini-batch with each iteration.

We used five-fold cross validation to evaluate the performance of
our model on the challenge training dataset, in which a few hype-parameters
were also experimentally determined via grid search. The total number
of iterations was set as $600$ for each fold. When applying the trained
models onto the testing dataset, a bagging-type ensemble strategy
was implemented to average the outputs of the six models to further
improve the segmentation performance. One iteration in model training
took about $60$ seconds using a single NVIDIA Geforce GTX 1060 GPU
with 1280 cores and 6GB memory. Applying the entire segmentation framework
on a new test image was, however, very efficient, taking about $0.2$
second for a typical $768\times1024$ image. 

\section{Experiments and results\label{sec:Experimental-design-and}}

\subsection{Evaluation metrics}

We applied the challenge evaluation metrics to evaluate the performance
of our method by comparing the computer-generated lesion masks with
the ground truths created by human experts. The evaluation metrics
include pixel-wise accuracy (AC), sensitivity (SE), specificity (SP),
dice coefficient (DI), and Jaccard index (JA):
\begin{itemize}
\item $AC=(TP+TN)/(TP+FP+TN+FN)$ 
\item $DI=2\cdot TP/(2\cdot TP+FN+FP)$
\item $JA=TP/(TP+FN+FP)$
\item $SE=TP/(TP+FN)$
\item $SP=TN/(TN+FP)$, 
\end{itemize}
where $TP$, $TN$, $FP$, $FN$ refer to the number of true positives,
true negatives, false positives, and false negatives respectively.
The final rank was based on $JA$ for this challenge.

\subsection{Experiments on network architectures}

In order to investigate if the increase of network depth with smaller
kernel size can improve the discriminative capability of CDNN and
thus yield a better segmentation performance, we compared the performance
of the proposed deep CDNN, which we denoted as CDNN-29, with the one
proposed in our previous work, which we denoted as CDNN-19. Since
only the results on the validation dataset could be obtained during
the challenge, this comparison was initially conducted on the validation
dataset, as shown in Table \ref{tab:depth-validation}. The ground
truth of testing dataset was held out by the organizer for final performance
evaluation during the challenge, and later released to the public
to encourage further investigations. So we extended this comparison
to the testing dataset in this experiment, as shown in Table \ref{tab:depth-testing}.

\begin{table}[tp]
\caption{Comparison of different architectures on validation dataset \label{tab:depth-validation}}
\centering{}%
\begin{tabular}{llccccc}
\hline 
Image size &  & AC & DI & JA & SE & SP\tabularnewline
\hline 
\hline 
CDNN-19 &  & $0.945$ & $0.839$ & $0.749$ & $0.854$ & $\boldsymbol{0.982}$\tabularnewline
CDNN-29 &  & $\boldsymbol{0.953}$ & $\boldsymbol{0.865}$ & $\boldsymbol{0.783}$ & $\boldsymbol{0.879}$ & $0.979$\tabularnewline
\hline 
\end{tabular}
\end{table}

\begin{table}[tp]
\caption{Comparison of different architectures on testing dataset \label{tab:depth-testing}}
\centering{}%
\begin{tabular}{llccccc}
\hline 
Image size &  & AC & DI & JA & SE & SP\tabularnewline
\hline 
\hline 
CDNN-19 &  & $0.921$ & $0.824$ & $0.736$ & $0.796$ & $\boldsymbol{0.979}$\tabularnewline
CDNN-29 &  & $\boldsymbol{0.934}$ & $\boldsymbol{0.849}$ & $\boldsymbol{0.765}$ & $\boldsymbol{0.825}$ & $0.975$\tabularnewline
\hline 
\end{tabular}
\end{table}

It is clear to see that the new CDNN model achieved better segmentation
performance in most of the metrics on both validation and testing
datasets. These results demonstrate that increasing network depth
while maintaining equivalent receptive field by reducing filter size
can effectively improve the discriminative capability of CDNN. 

\subsection{Experiments on input channels}

\begin{figure}[tp]
\begin{centering}
\textsf{\includegraphics[scale=1.6]{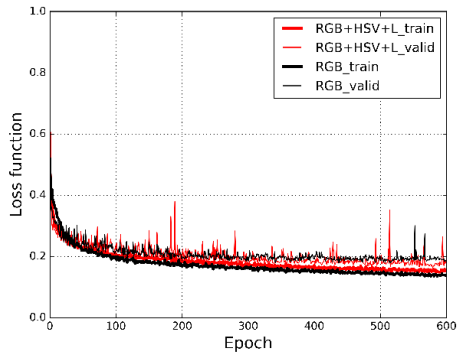}}
\par\end{centering}
\caption{Comparison of network inputs with and without the additional HSV+L
color channels for training and validating of the proposed model.\label{fig:channel-comparison}}
\end{figure}

Using CDNN-29 model, we evaluated how the additional HSV and L channels
affect skin lesion segmentation performance. Figure \ref{fig:channel-comparison}
shows how the loss function is minimized as network training evolves.
The black lines represent the model training and validation using
RGB channels, and the red lines add additional four channels as model
input. It is clear to see that the gap between training and validation
errors was effectively narrowed down when including the additional
four color channels as network input. Meanwhile, the validation error
with additional color channels was consistently lower than that of
RGB alone, yielding improved performance on the new testing images.
As shown in Table \ref{tab:channel-validation} and \ref{tab:channel-testing},
the additional four input channels improved the Jaccard index by $2.4\%$
and $1.5\%$ on validation and testing datasets, respectively.

\begin{table}[tp]
\caption{Comparison of different input channels on validation dataset \label{tab:channel-validation}}
\centering{}%
\begin{tabular}{llccccc}
\hline 
Image size &  & AC & DI & JA & SE & SP\tabularnewline
\hline 
\hline 
RGB &  & $0.951$ & $0.848$ & $0.765$ & $0.868$ & $0.976$\tabularnewline
RGB+HSV+L &  & $\boldsymbol{0.953}$ & $\boldsymbol{0.865}$ & $\boldsymbol{0.783}$ & $\boldsymbol{0.879}$ & $\boldsymbol{0.979}$\tabularnewline
\hline 
\end{tabular}
\end{table}

\begin{table}[tp]
\caption{Comparison of different input channels on testing dataset \label{tab:channel-testing}}
\centering{}%
\begin{tabular}{llccccc}
\hline 
Image size &  & AC & DI & JA & SE & SP\tabularnewline
\hline 
\hline 
RGB &  & $0.931$ & $0.840$ & $0.754$ & $0.813$ & $\boldsymbol{0.976}$\tabularnewline
RGB+HSV+L &  & $\boldsymbol{0.934}$ & $\boldsymbol{0.849}$ & $\boldsymbol{0.765}$ & $\boldsymbol{0.825}$ & $0.975$\tabularnewline
\hline 
\end{tabular}
\end{table}

\subsection{Comparison with other methods in the challenge}

During the 2017 ISBI challenge on skin lesion segmentation, $39$
teams evaluated their algorithms during validation phase, but only
$21$ teams were able to participate the final official challenge
by submitting their results on the $600$ testing images. Table \ref{tab:Table-isbi-testing}
lists the results from the top ten teams. Because the previous studies
have shown the advantage of using CNNs in skin lesion segmentation
over other traditional methods \cite{yu2017automated,yuan2017automatic},
most of the top teams in this challenge included various deep neural
network models in their segmentation methods, with the exception that
the NedMos team employed active contour on the saliency map \cite{jahanifar2017seg}.
While each team was allowed for multiple submissions, only the most
recent one would be considered to determine the team's final score.
Our method achieved an average Jaccard Index (JA) of $0.765$, ranking
as the first place in this challenge. 

\begin{table}[tp]
\caption{Results of 2017 ISBI challenge on skin lesion segmentation\label{tab:Table-isbi-testing}}
\centering{}%
\begin{tabular}{lccccc}
\hline 
Team & AC & DI & JA & SE & SP\tabularnewline
\hline 
\hline 
MtSinai (ours) & $\boldsymbol{0.934}$ & $\boldsymbol{0.849}$ & $\boldsymbol{0.765}$ & $\boldsymbol{0.825}$ & $0.975$\tabularnewline
NLP LOGIX & $0.932$ & $0.847$ & $0.762$ & $0.820$ & $0.978$\tabularnewline
USYD (Bi) & $\boldsymbol{0.934}$ & $0.844$ & $0.760$ & $0.802$ & $\boldsymbol{0.985}$\tabularnewline
USYD (Ann) & $\boldsymbol{0.934}$ & $0.842$ & $0.758$ & $0.801$ & $0.984$\tabularnewline
RECOD & $0.931$ & $0.839$ & $0.754$ & $0.817$ & $0.970$\tabularnewline
Jer & $0.930$ & $0.837$ & $0.752$ & $0.813$ & $0.976$\tabularnewline
NedMos & $0.930$ & $0.839$ & $0.749$ & $0.810$ & $0.981$\tabularnewline
INESC & $0.922$ & $0.824$ & $0.735$ & $0.813$ & $0.968$\tabularnewline
Shenzhen U (Lee) & $0.922$ & $0.810$ & $0.718$ & $0.789$ & $0.975$\tabularnewline
GAMMA & $0.915$ & $0.797$ & $0.715$ & $0.774$ & $0.970$\tabularnewline
\hline 
\end{tabular}
\end{table}

\begin{figure*}[tbph]
\begin{centering}
\textsf{\includegraphics[scale=0.8]{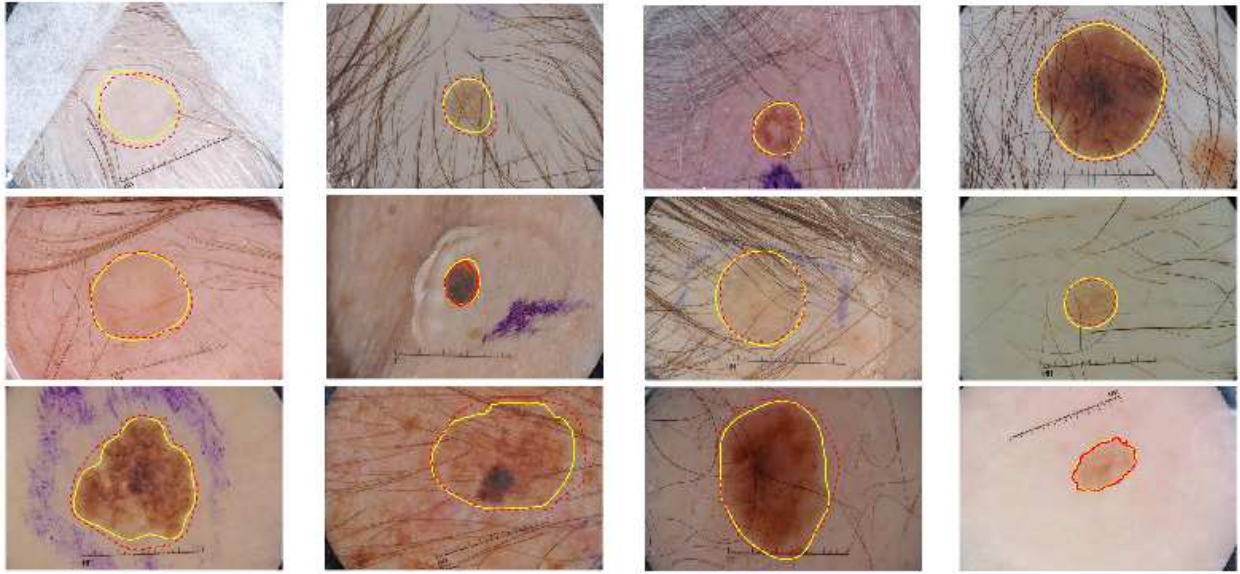}}
\par\end{centering}
\caption{Examples of automatic segmentation on the ISBI 2017 challenge testing
dataset, including four nevus lesions in the top row, four seborrheic
keratosis lesions in the middle rows and four melanoma lesions in
the bottom row. In each figure, the red dash line indicates the outline
contoured by dermatologist and the yellow solid line is the result
of automatic segmentation.\label{fig:isbi-examples}}
\end{figure*}

Figure \ref{fig:isbi-examples} shows a few challenging examples of
automatic skin lesion segmentation with our CDNN model, demonstrating
the robustness of the proposed model under various imaging acquisition
conditions.

\section{Discussions\label{sec:Discussions}}

While deep convolutional neural networks have shown remarkable success
in various medical image segmentation tasks, it is still a challenging
task to extend those models into a scaled-up study where a large amount
of image data are involved. In this study, we investigated if a deeper
CDNN model coupled with additional input information is capable of
handling the increased complexity introduced from the much dynamic
appearance of the targeting objects. It should be noted that deep
neural networks have a very flexible architecture and different network
design certainly matters. However, there is no clear guideline about
what the optimal network architecture would be for a given application
and available data. While it is infeasible to explore all the possible
options in our study due to the long training time, we participated
the ISBI 2017 challenge on skin lesion segmentation, which not only
provides a large database to evaluate the performance of our model,
but also allows researchers to compare different designs using the
same benchmark data.

All of the top five teams employed deep learning in their segmentation
methods, demonstrating the popularity and effectiveness of FCN-based
methods in medical image segmentation. These methods cover a large
spectrum of FCN models and training strategies. For example, Berseth
from NLP LOGIX \cite{berseth2017isic} employed a U-Net architecture
\cite{ronneberger2015u} and tried to include a conditional random
field (CRF) \cite{chen2014semantic} as post-processing in their method.
Besides the $2000$ training images provided by the challenge organizers,
Bi and Ann from USYD \cite{bi2017auto} trained their deep residual
networks with additional $\sim8000$ images from ISIC archive. We
attribute the superior performance of our CDNN model to the following
three aspects: 1) A deeper architecture with small kernel allowed
more non-linear transformations on the data while reducing the number
of trainable parameters as compared to its shallower counterpart,
yielding significantly improved discriminant capability; 2) Additional
input features made network training more efficient and robust by
explicitly including complementary but useful information from other
color spaces; 3) The loss function based on Jaccard distance enabled
the network training to naturally focus more on lesion pixels over
background, which further improved the segmentation performance by
lifting sensitivity (SE), as demonstrated in Table \ref{tab:Table-isbi-testing}.

\begin{figure}[tp]
\begin{centering}
\textsf{\includegraphics[scale=0.55]{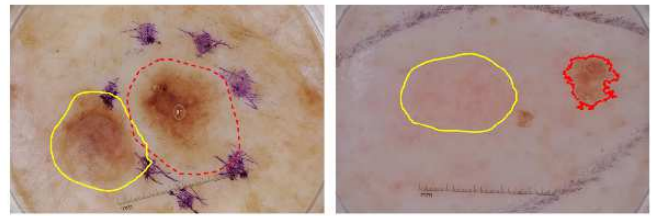}}
\par\end{centering}
\caption{Cases where our model failed due to mis-segmentation to the wrong
sites. The red dash and yellow solid contours indicate the ground
truth and the segmentation results, respectively.\label{fig:failed-case}}
\end{figure}

\begin{figure}[tp]
\begin{centering}
\textsf{\includegraphics[scale=2]{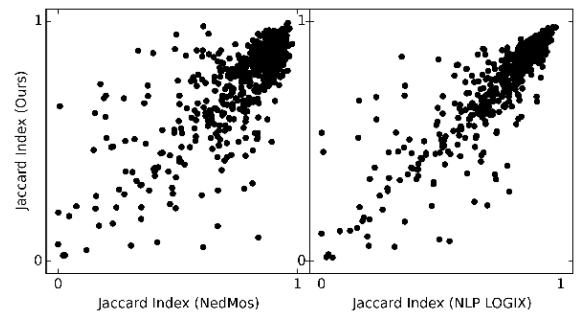}}
\par\end{centering}
\caption{Scatter plots of our model vesus NedMos (left), and NLP LOGIX (right)
.\label{fig:scatter-plots}}
\end{figure}

The segmentation performance on some cases is still rather low, as
shown in two examples in Figure \ref{fig:failed-case}. Thus, further
improvement is certainly needed. In additional to better network architecture
and more effective training strategies, one possible way is to combine
the deep learning models with the conventional image segmentation
methods, such as active contour. For example, we found the correlation
between the outputs of our model and MedMos (active contour model)
was $0.759$, while it was $0.873$ with NLP LOGIX (another deep FCN
model). The scatter plots are shown as Figure \ref{fig:scatter-plots}.
Since the active contour is less correlated with our model, it may
provide more complementary information when ensembling multiple models
for further segmentation. Another option is to integrate other post-processing
techniques, such as CRF, into our CDNN model. We actually attempted
to employ this technique in this challenge, but ultimately discard
it due to the inferior performance. One possible reason is that the
parameters to determine the unary and pariwise potentials were pre-set
and thus lack of flexibility to handle the large variations of the
data. A more dynamic mechanism, such as plugging CRF in as a part
of FCN by modeling it as recurrent neural networks (RNNs) \cite{zheng2015conditional},
may resolve this issue and thus improve the segmentation performance.
These are all interesting research topics and worthy of further investigations.

\section{Conclusion\label{sec:Conclusion}}

In this paper, we present a major extension of our previously proposed
CDNN model in automatic skin lesion segmentation on dermoscopic images.
Our new model leverages the increased discriminant capability of deeper
network structures with smaller convolutional kernels to segment skin
lesions in a much larger variety of image acquisition conditions.
The segmentation performance was further boosted by combining information
from multiple color spaces. Our approach excelled other state-of-the-art
methods when evaluating on ISBI 2017 challenge of skin lesion segmentation.
Our network architectures and training strategies are inherently general
and can be easily extended to other applications.

\section{Acknowledgment}

The authors are grateful to the organizers of 2017 International Symposium
on Biomedical Imaging (ISBI 2017) challenge of $Skin$ $Lesion$ $Analysis$
$Towards$ $Melanoma$ $Detection$ for organizing such an inspiring
challenge.

\end{document}